\documentclass[10pt,journal,compsoc]{IEEEtran}
\ifCLASSOPTIONcompsoc
  \usepackage[nocompress]{cite}
  \usepackage{amsthm,amsmath,amssymb}
  \usepackage{mathrsfs}
  \usepackage{graphicx,epstopdf}
  \usepackage{multirow}
  \usepackage{subfigure}
  \usepackage{caption2}
  \usepackage[noend]{algpseudocode}
  \usepackage{algorithmicx,algorithm}
  \usepackage{makecell}
\else
  \usepackage{cite}
\fi
\ifCLASSINFOpdf
\else
\fi
\begin{document}
%
\title{Exploring Uncertainty in Deep Learning for Construction of Prediction Intervals}

\author{Yuandu~Lai, 
        Yucheng~Shi,
        Yahong~Han,~\IEEEmembership{Member,~IEEE,}
        Yunfeng~Shao,
        Meiyu~Qi,
        Bingshuai~Li



\thanks{Y. Lai, Y. Shi and Y. Han are with College of Intelligence and Computing,
Tianjin University, Tianjin, China (e-mail: \{yuandulai, yucheng, yahong\}@tju.edu.cn).}
\thanks{Y. Shao, M. Qi and B. Li are with Huawei Noah's Ark Lab, Huawei Technologies (e-mail: \{shaoyunfeng, qimeiyu, libingshuai\}@huawei.com).}
}

%
%

\markboth{MANUSCRIPT SUBMITTED TO IEEE TRANSACTIONS ON KNOWLEDGE AND DATA ENGINEERING}
{Shell \MakeLowercase{\textit{et al.}}: Bare Demo of IEEEtran.cls for Computer Society Journals}
%



\IEEEtitleabstractindextext{%
\begin{abstract}
Deep learning has achieved impressive performance on many tasks in recent years. However, it has been found that it is still not enough for deep neural networks to provide only point estimates. For high-risk tasks, we need to assess the reliability of the model predictions. This requires us to quantify the uncertainty of model prediction and construct prediction intervals. In this paper, We explore the uncertainty in deep learning to construct the prediction intervals. In general, We comprehensively consider two categories of uncertainties: aleatory uncertainty and epistemic uncertainty. We design a special loss function, which enables us to learn uncertainty without uncertainty label. We only need to supervise the learning of regression task. We learn the aleatory uncertainty implicitly from the loss function. And that epistemic uncertainty is accounted for in ensembled form. Our method correlates the construction of prediction intervals with the uncertainty estimation. Impressive results on some publicly available datasets show that the performance of our method is competitive with other state-of-the-art methods.
\end{abstract}

}

\maketitle

\IEEEdisplaynontitleabstractindextext

%
\IEEEpeerreviewmaketitle

\IEEEraisesectionheading{\section{Introduction}\label{sec:introduction}}

%
%
%
%
\IEEEPARstart{W}{ith} the rapid development of artificial intelligence, deep learning has attracted the interest of many researchers~\cite{lecun2015deep}. As a representative tool of deep learning, deep neural networks has achieved impressive performance in many tasks. At present, deep neural networks have important applications in computer vision~\cite{krizhevsky2012imagenet}, speech recognition~\cite{hinton2012deep}, natural language processing~\cite{mikolov2013efficient}, bioinformatics~\cite{zhou2015predicting} and other fields. Although deep neural networks have achieved high accuracy for many tasks, they still perform poorly in quantifying the uncertainty of predictions and tend to be overconfident. For many real-world applications, it is not enough for a model to be accurate in its predictions. It must also be able to quantify the uncertainty of each prediction. This is especially importance in tasks where wrong prediction has a great negative impact, such as: machine learning auxiliary medical diagnosis~\cite{esteva2017dermatologist}, automatic driving~\cite{bojarski2016end}, fiance and energy system, etc. These high-risk applications require not only point prediction, but also the precise quantification of the uncertainty. So it can be said that in many application scenarios, uncertainty is very important, more important than precision. Moreover, uncertainty plays an important role in determining when to abandon the prediction of the model. Abandoning the inaccurate predictions of the model can handle exceptions and hand over high-risk decisions to humans~\cite{geifman2017selective}. As more and more autonomous systems based on deep learning are deployed in our real life that may cause physical or economic harm, we need to better understand when we can have confidence in the prediction of deep neural networks and when we should not be so sure.

The development of neural networks precise prediction intervals (PIs)~\cite{heskes1997practical} is a challenging work to explore the uncertainty of prediction, which has just aroused the interest of researchers. PIs directly conveys uncertainty, providing a lower bound and an upper bound for prediction\cite{pearce2018high}, but traditional deep neural networks can only provide a point estimation. PIs is of great use in practical applications, and can help people make better decisions. However, although there are some researches on uncertainty estimation~\cite{blundell2015weight}~\cite{lakshminarayanan2017simple}~\cite{gal2016dropout}~\cite{malinin2018predictive}~\cite{liu2001new}, there are few researches on PIs. In this paper, We will
explore the uncertainty in deep learning to construct the prediction intervals.

Uncertainty comes from different sources in various forms, and most literature considers two sources of uncertainty: aleatory uncertainty and epistemic uncertainty~\cite{der2009aleatory}~\cite{hoffman1994propagation}~\cite{kendall2017uncertainties}~\cite{detlefsen2019reliable}~\cite{depeweg2017decomposition}~\cite{li2012dealing}. Aleatory uncertainty describes the irreducible inherent noise of the observed data. This is due to the complexity of the data itself, which can be sensor noise, label noise, class overlap, etc. Aleatory uncertainty is also caused by hidden variables that are not observed or errors in the measurement process, and this type of uncertainty cannot be reduced by collecting more data. Aleatory uncertainty is also known in some literature as data uncertainty, random uncertainty~\cite{henrion1986assessing}, stochastic uncertainty~\cite{helton1994treatment}. Epistemic uncertainty describes the model errors due to lack of experience in some regions of the feature space\cite{tagasovska2019single}. In some feature spaces, if the training samples do not cover them, the uncertainty will increase in those places. Therefore, the epistemic uncertainty is inversely proportional to the density of the training samples, which can be reduced by collecting more data in the low-density areas. In regions with high-density training samples, epistemic uncertainty will decrease, and aleatory uncertainty caused by data noise will play a major role. Epistemic uncertainty is also known as model uncertainty~\cite{detlefsen2019reliable}, because it is caused by the limitations of the model.

By analyzing the sources and categories of uncertainty,  we can deal with and estimate each kind of uncertainty well in practice.


The precise prediction intervals of neural network is a difficult task to solve, and its development is closely related to uncertainty estimation. The prediction interval directly reflects the uncertainty. Many literatures have proposed that the loss function can be modified to implicit learning prediction intervals~\cite{heskes1997practical}~\cite{pearce2018high}~\cite{lakshminarayanan2017simple}~\cite{keren2018calibrated}~\cite{kuleshov2018accurate}. These methods have obvious disadvantages, such as loss function can not be optimized by gradient descent-based algorithm, which is a modern machine learning technology. Or the accuracy of PIs is low. Although there are many confidence interval construction methods with good performance in statistical literature, these methods can not be effectively combined with deep learning.

In this paper, we propose a prediction interval construction method from the perspective of statistics which can be used effectively only by slight modification of deep neural networks. We modify the last layer of the general deep neural network so that the network no longer outputs only a point estimation, but the upper bound and lower bound of the interval. And a loss function that can be used to train such a network is derived from the perspective of probability theory. Because most modern deep neural networks can use our loss function with only a few modifications, our method will have more practical use than other methods.

The contributions of this work are summarized as follows: i) In this paper, the construction of optimal prediction intervals is associated with uncertainty estimation. We design a novel loss function based on the theory of statistics. We think our work bridges the gap between the construction of prediction intervals and the uncertainty estimation; ii) Unlike bayesian deep learning, our loss function can be seamlessly integrated with current deep learning frameworks such as $TensorFlow$ and $PyTorch$. Existing deep learning networks can use our loss function with only minor modifications. And it can be optimized efficiently using standard optimization techniques. We believe that our method promotes the development of non-bayesian method to construct prediction interval. iii) Experimental results show that this method can construct compact optimal prediction intervals. It can sensitively capture the epistemic uncertainty. In the low-density data distribution area, the uncertainty increases obviously. Experimental results on some publicly available regression datasets show the effectiveness of the method.

The rest of the paper is organized as follows. In the section 2, we summarize the related work of uncertainty estimation and prediction interval. And we introduce the method of constructing prediction interval based on exploring uncertainty in section 3. In the section 4, we organize some convincing experiments to show the effectiveness of our method. Finally, in the section 5, we make a brief summary of this paper.

\section{RELATED WORK}
Although it is impossible to eliminate the uncertainty completely, it is worthwhile to reasonably estimate and deal with the uncertainty to avoid the cost caused by the error of high-risk prediction. In this section, current research in this field will be presented.

\textbf{Gaussian processes (GPs).}
  GPs has a simple structure, and the algorithm model is determined by the mean function and the covariance function~\cite{nix1994estimating}. And GPs is a good method of function approximation because of its flexibility, robustness to overfitting and provide well-calibrated aleatory uncertainty~\cite{salimbeni2017doubly}~\cite{rasmussen2003gaussian}. However, the gaussian process also has obvious disadvantages. It is not sparse, which makes it difficult to process high-dimensional data or massive data. And its prediction accuracy is not as good as that of deep neural networks. Recent research on GPs, such as sparse GPs~\cite{snelson2006sparse} and deep GPs~\cite{damianou2013deep}, is addressing these problems.

  \textbf{Bayesian neural networks (BNN)~\cite{mackay1992practical}.}
  In bayesian estimation, the prior distribution hypothesis of network parameters is introduced, and the posterior distribution is calculated by using bayesian formula, so as to introduce the uncertainty into the prediction of neural network. In practice, in order to introduce bayesian estimation into neural network, the method of variational inference~\cite{hoffman2013stochastic} is often used to approximate the posterior distribution. The latest literature on BNN is available~\cite{graves2011practical}~\cite{louizos2016structured}~\cite{kingma2013auto}~\cite{louizos2017multiplicative}~\cite{rezende2014stochastic}. BNN can be simply understood as introducing uncertainty into the weight of neural network, which is equivalent to integrating infinite neural networks on a certain weight distribution for prediction. However, this kind of method has obvious disadvantages, such as high computational complexity and the need for distribution hypothesis.

 \textbf{Ensemble methods.}
  Estimating epistemic uncertainty is learning which areas of the feature space are unexplored by the training samples. Most estimation of epistemic uncertainty are based on measuring variance between multiple models. The prediction results of multiple models trained on the training set are integrated into the final prediction value of the model and the estimate of epistemic uncertainty. Most recent approaches to estimating epistemic uncertainty follow this principle. Lakshminarayanan \emph{et al.}~\cite{lakshminarayanan2017simple} proposed a simple and scalable non-bayesian solution that provides a very good baseline for evaluating the uncertainty. The authors train a probabilistic neural network to estimate the aleatory uncertainty using the appropriate loss function. In addition, the author also used the idea of ensemble learning~\cite{dietterich2002ensemble} to boost predictive performance, which is an important work of non-bayesian method in predicting uncertainty.

  In recent years, a lot of work has been done on the development of approximate bayesian inference methods. Some works~\cite{blundell2015weight}~\cite{hernandez2015probabilistic} place an independent gaussian priori on each weight in the neural network, and then use back propagation to learn the mean and variance of these gaussian. After training, the weight distribution can be used to sample different networks to obtain different predictions.

  Gal \emph{et al.}~\cite{gal2016dropout} proposed a new theoretical framework, which uses the dropout strategy~\cite{srivastava2014dropout} in deep neural networks as approximate bayesian inference. Bayesian probability theory provides us with a math-based method to infer the uncertainty of models, but it usually comes with a high computational cost. In this paper, the authors demonstrate that it is possible to convert recent deep learning tools to a bayesian model without changing the model. The trained neural network model is tested for many times, and Monte Carlo dropout is carried out every time. The model uncertainty can be quantified by calculating the entropy or variance of the predictions of the model. However, this approach has obvious limitation on the network model structure, which must include the dropout layer.

  Since dropout has been largely replaced by batch normalization~\cite{ioffe2015batch}~\cite{he2016identity}, Teye \emph{et al.}~\cite{teye2018bayesian} demonstrate how to use the Monte Carlo batch normalization (MCBN) to make meaningful estimation of model uncertainty. Further, this discovery enables us to make appropriate estimation of model uncertainty for traditional network architectures without modifying the network or training process. Although these methods avoid the disadvantages of bayesian methods, they can only measure one uncertainty, which is not complete.

\textbf{Prediction Intervals.}
  Prediction intervals is a widely used tool to quantify the uncertainty of predictions. PIs directly conveys uncertainty. A representative work is proposed by Pearce \emph{et al.}~\cite{pearce2018high}, which is based on the improvement of khosravi \emph{et al.}~\cite{khosravi2010lower}. The author thinks that high quality PIs should be as narrow as possible and cover the data points of specified proportion. The author calls it the high quality criterion. According to this criterion, the author deduces a loss function, which is used to predict PIs without distribution assumption. In addition, a common method for prediction of interval range is quantile regression~\cite{koenker2001quantile}. It was first proposed by Koenker \emph{et al.}~\cite{koenker1978regression}. On the other hand, Rosenfeld \emph{et al.} put forward the method of building PIs in the setting of batch learning, which is called $IntPred$~\cite{rosenfeld2018discriminative}, that is, building PIs of a group of test points at the same time. An obvious disadvantage of these methods is that the PIs they constructed have nothing to do with uncertainty or have little relevance.

 \textbf{Applications of uncertainty.}
  Firstly, uncertainty plays an important role in deciding when to abandon the prediction of the model and let humans make decisions to avoid the loss caused by mistakes~\cite{geifman2017selective}. Secondly, the study of uncertainty can promote the development of active learning~\cite{8737749}, that is, deciding which data people should label to maximize the performance of the model~\cite{settles2009active}. Thirdly, uncertainty estimation is very important when constructing the prediction intervals~\cite{keren2018calibrated}.

\section{Construction of Optimal Prediction Intervals Based on Uncertainty Estimation}
In this paper, we propose a method of constructing prediction intervals based on quantifying two kind of uncertainty in deep neural networks. We call our method uncertainty-based PI (UBPI).
\subsection{Notation and Problem Setup}
Let $\boldsymbol{x} \in \mathbb{R}^d$ represent the d-dimensional features. For the regression problem, the label is assumed to be real-valued, that is $y \in \mathbb{R}$. And we assume that a training dataset $\mathcal{D}$ consists of $N$ i.i.d. samples $\mathcal{D}=\{(\boldsymbol{x_n}, y_n)\}^N_{n=1}$.

In order to explain the uncertainty in the regression scenario conveniently, let's consider the following formula:
\begin{equation}\label{eq1}
y = f(\boldsymbol{x}) + \sigma_{noise},
\end{equation}
where statistically $f(\cdot)$ is some data generating function, and $\sigma_{noise}$ is the irreducible noise, also named the data noise. The reason it exists is because there are unobserved explanatory variables or due to the inherent randomness in nature. In general, the goal of regression is to produce an estimate $\hat{f}(x)$ that can be used to make a prediction of point estimation. The two terms in the eq.(\ref{eq1}) correspond to the sources of epistemic uncertainty and aleatory uncertainty respectively. Therefore, when we estimate the uncertainty of $y$, assuming that the two uncertainties are independent, then the total uncertainty is:
\begin{equation}\label{eq2}
\sigma^2 = \sigma^2_{ep} +\sigma^2_{al},
\end{equation}
where $\sigma^2_{ep}$ denotes the epistemic uncertainty, $\sigma^2_{al}$ denotes the aleatory uncertainty.

\subsection{Construct the Optimal Prediction Interval}
Now we introduce how to construct optimal prediction intervals without the 'uncertainty labels'. We assume that the size of a mini-batch is $n$, and with $\boldsymbol{x_i}$ denoting the $i$th input corresponding to $y_i$. The philosophy of uncertainty quantification is to predict a prediction interval (PI) [$\hat{y}_{Li}, \hat{y}_{Ui}$] to bound $y_i$ to satisfy:
\begin{equation}\label{eq3}
  Pr[\hat{y}_{Li} \leq y_i \leq \hat{y}_{Ui}] \geq P_c,
\end{equation}
where $P_c$ is the predefined confidence level.

\subsubsection{Hybrid Loss Function for Prediction Intervals}
In order to equip UBPI with the capability of quantifying uncertainty and providing optimal prediction intervals, the training loss function $\mathcal{L}$ has two terms:
\begin{equation}
  \mathcal{L} = \mathcal{L}_{UE} + \lambda\mathcal{L}_{PI},
\end{equation}
where hyper-parameter $\lambda$ balances the importance of two terms.

\begin{figure*}
\centering
\includegraphics[width=1.0\textwidth]{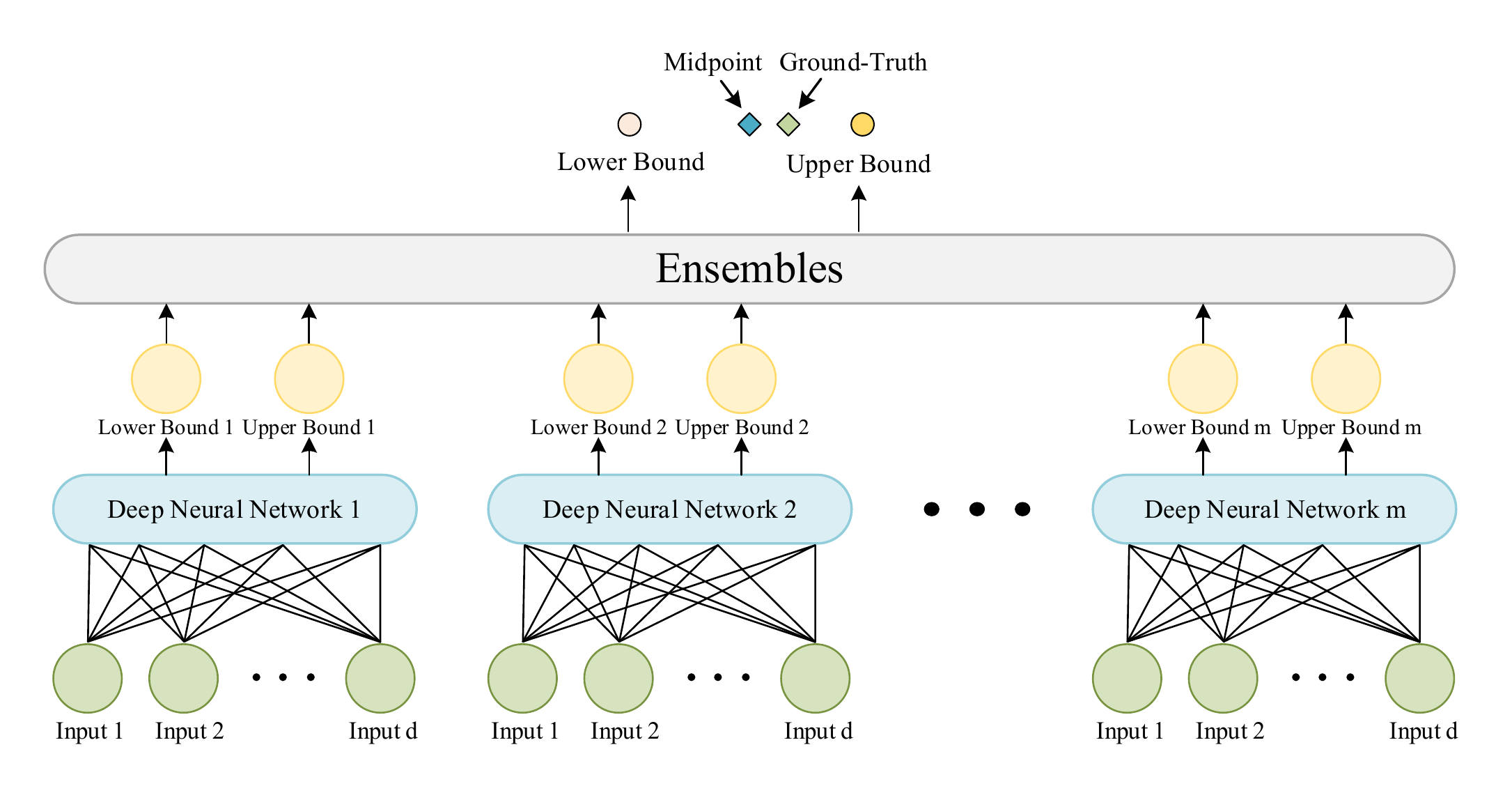}
\caption{Ensembles of neural networks. A group of neural networks are trained with different parameter initializations. $m$ is the size of the ensemble of neural networks. The resulting ensemble of neural networks will contain some diversity. The variance of their predictions is interpreted as an estimate of epistemic uncertainty.}
\label{fig3}
\end{figure*}

We now derive $\mathcal{L}_{UE}$ from estimating the aleatory uncertainty in the regression task. For regression task, we often define model likelihood as a Gaussian with mean given by the model output and an observation noise scalar $\sigma$: $\mathcal{N}(\hat{f}(x), \sigma^2)$. We fix this Gaussian likelihood to model aleatoric uncertainty in regression tasks. For a regression problem, its likelihood is the product of the corresponding value of the probability density function (PDF). Combined with the PDF of Gaussian distribution, the negative log likelihood (NLL)~\cite{quinonero2005evaluating} of regression can be derived as follows:
\begin{equation}
  NLL = -\sum_{i=1}^{n}log[\frac{1}{\sqrt{2\pi}\sigma}exp(-\frac{(y_i-\hat{y}_{i})^2}{2\sigma^2})]
\end{equation}

\begin{equation}\label{nll}
  = \frac{n}{2}[\frac{\frac{1}{n}\sum_{i=1}^{n}(y_i-\hat{y}_{i})^2}{\sigma^2} + log(\sigma^2)  + log(2\pi)],
\end{equation}
with $\sigma^2$ capturing how much uncertainty we have in the outputs.

Here, we use the width of the interval as an estimate of the uncertainty. In practice, we simplify eq.(\ref{nll}), and $\mathcal{L}_{UE}$ is designed in the following form:
\begin{equation}\label{eq5}
  \mathcal{L}_{UE} = \frac{n}{2}[\frac{MSE}{MPIW} + log MPIW],
\end{equation}
where MSE is Mean Squared Error of a mini-batch, and we use the midpoint of the interval [$\hat{y}_{Li}, \hat{y}_{Ui}$] as the point estimation:
\begin{equation}
  \hat{y}_{i}= \frac{\hat{y}_{Li} + \hat{y}_{Ui}}{2},
\end{equation}
\begin{equation}
  MSE = \frac{1}{n}\sum_{i=1}^{n}(y_i-\hat{y}_{i})^2.
\end{equation}
And $MPIW$ is Mean Prediction Interval Width ~\cite{pearce2018high}~\cite{khosravi2010lower} ~\cite{khosravi2010construction}, which is defined as:
\begin{equation}\label{eq8}
  MPIW = \frac{1}{n}\sum_{i=1}^{n}(\hat{y}_{Ui} - \hat{y}_{Li}).
\end{equation}

We have replaced the $\sigma^2$ in eq.(\ref{nll}) with MPIW. We think the interval width is a good indicator of the uncertainty. That is to say, the wider the interval, the greater the uncertainty. On the contrary, the narrower the interval, the smaller the uncertainty. The $\mathcal{L}_{UE}$ loss consists of two components: a residual regression term and an uncertainty regularization term. We do not need uncertainty labels to learn uncertainty. Instead, we only need to supervise the learning of the regression task. We learn the  aleatory uncertainty implicitly from the loss function. The second regularization term prevents the network from generating too wide prediction intervals, which is meaningless.

To learn prediction interval, $\mathcal{L}_{PI}$ is inspired from the study~\cite{pearce2018high}~\cite{khosravi2010lower} and constructed as below:
\begin{equation}\label{eq9}
  \mathcal{L}_{PI} = max(0, P_c - PICP)^2,
\end{equation}
where $PICP$ is PI coverage probability. $PICP$ is the spontaneous measure related to the quality of the constructed PIs, and is denoted as:
\begin{equation}\label{eq10}
  PICP = \frac{1}{n}\sum_{i=1}^{n}c_i,
\end{equation}
where $c_i$ = 1 if $y_i \in [\hat{y}_{Li}, \hat{y}_{Ui}]$, otherwise $c_i$ = 0. The advanced intuition of eq.(\ref{eq9}) is that if the $PICP$ is lower than the predefined confidence level $P_c$ during the training, penalty loss will be imposed.

\renewcommand{\algorithmicrequire}{\textbf{Input:}}  
\renewcommand{\algorithmicensure}{\textbf{Output:}} 
\begin{algorithm}[t]
\caption{Construction of hybrid loss function} 

\begin{algorithmic}
\Require
      Predictions of lower bound $\bf{\hat{y}_L}$ and upper bound $\bf{\hat{y}_U}$, ground truth label $\textbf{y}$, predefined confidence level $P_c$, hyper-parameter factor for softing $s$, hyper-parameter $\lambda$ balances the importance of two terms, n denotes the size of a mini-batch, $\sigma$ denotes sigmoid activation function, $\odot$ denotes the element-wise product.
\Ensure
      Loss
\State $\mathbf{\hat{y}} = (\mathbf{\hat{y}_L} + \mathbf{\hat{y}_U}) / 2$ 
\State $MSE = reduce\_mean(\mathbf{y} - \mathbf{\hat{y}})^2$
~\\
\State $MPIW = reduce\_mean(\mathbf{\hat{y}_U} - \mathbf{\hat{y}_L})$
\State $\mathbf{c} = \sigma(s\cdot(\mathbf{\hat{y}_{U}} - \mathbf{y}))\odot\sigma(s\cdot(\mathbf{y} - \mathbf{\hat{y}_{L}}))$
\State $PICP = reduce\_mean(\mathbf{c})$
~\\
\State $\mathcal{L}_{UE} = \frac{n}{2}[\frac{MSE}{MPIW} + log MPIW]$
\State $\mathcal{L}_{PI} = max(0, P_c - PICP)^2$
\State $\mathcal{L} = \mathcal{L}_{UE} + \lambda\mathcal{L}_{PI}$
\end{algorithmic}
\end{algorithm}

Due to the existence of operation $max(\cdot)$ in eq.(\ref{eq9}), applying the boolean $c_i$ in eq.(\ref{eq10}) can cause the $\mathcal{L}_{PI}$ loss not to be differentiable for back-propagation algorithm. So, following the previous work~\cite{pearce2018high}~\cite{wang2020deeppipe}~\cite{yan2003optimizing}~\cite{yan2004predicting}, we also replace $c_i$ with a softened version as follows:
\begin{equation}\label{eq11}
  c_i = \sigma(s\cdot(\hat{y}_{Ui} - y_i))\cdot\sigma(s\cdot(y_i - \hat{y}_{Li})),
\end{equation}
where $\sigma$ is the sigmoid activation function and $s$ is a hyper-parameter factor for softing. Algorithm 1. shows the process of constructing our hybrid loss function.

\begin{figure}[t]
\centering
\includegraphics[width=0.45\textwidth]{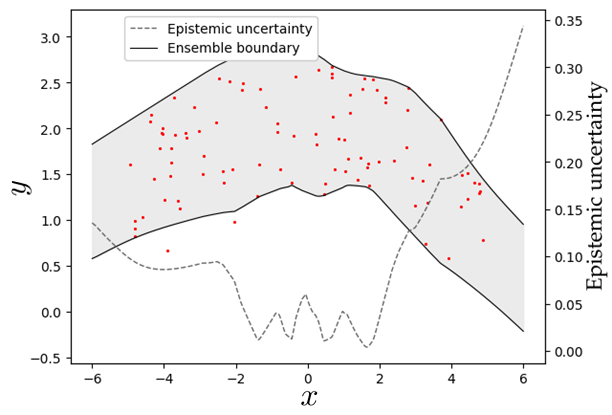}
\caption{Results on a toy regression task. The red dots are observed noisy training data points and the gray shade correspond to the 95\% prediction interval. This figure shows the compact PIs.}
\label{fig1}
\end{figure}

\subsubsection{Comparison of Various Loss Functions for Uncertainty Prediction}
In this section, we compare our loss function with two other loss functions that have been proposed to predict PIs. The key idea of these methods is to construct the PIs by minimizing the loss function based on eq.(\ref{eq3}) and eq.(\ref{eq8}). These loss functions typically contain two terms: $MPIW$ and $PICP$. This phenomenon shows that the recent work is to find the basic balance between coverage probability and PI width~\cite{khosravi2010lower}. For example, the loss function proposed by Khosravi \emph{et al.}~\cite{khosravi2010lower} is:
\begin{equation}
  Loss_{LUBE} = \frac{MPIW}{r}(1 + exp(\lambda max(0,(1 - \alpha) - PICP))),
\end{equation}
where $r$ is the numerical range of the target variable $y$, and $\lambda$ controls the importance between $PICP$ and $MPIW$. And another loss function proposed by hu \emph{et al.}~\cite{hu2019mbpep} is called $MBPEP$ and its form is as follows:
\begin{equation}
   Loss_{MBPEP} = MPIW + \lambda\cdot ReLU((1-\alpha)-PICP),
\end{equation}
where $(1 - \alpha)$ is the predefined confidence level and $ReLU(\cdot)$ denotes the Rectified Linear Unit (ReLU).

The proposed loss function differs from the existing work of PIs estimation in many aspects. Firstly, where our loss function differs from them is in the use of the term $MPIW$. $Loss_{MBPEP}$ adds $MPIW$ as a single term to the loss function, while we integrate $MPIW$ into the form of NLL as an estimate of uncertainty. That is, our loss function bridges the gap between the width of the PIs and the uncertainty. And intuitively, $MBPEP$ loss function is not strongly related to uncertainty estimation. Secondly, ours loss function is differentiable and can be effectively minimized using standard optimization techniques like Stochastic Gradient Descent (SGD). However, the $Loss_{LUBE}$ loss function is not differentiable, which makes it difficult to optimize.

\begin{figure}[t]
\centering
\includegraphics[width=0.43\textwidth]{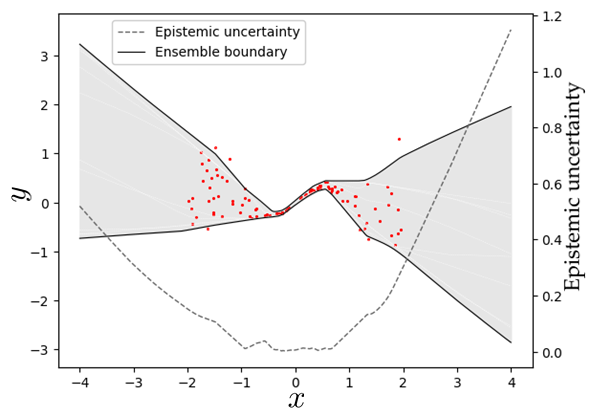}
\caption{Results on a toy regression task. The red dots are observed noisy training data points and the gray shade correspond to the 95\% prediction interval. This figure shows  capturing epistemic uncertainty.}
\label{fig2}
\end{figure}

\begin{table}[t]
\caption{The UCI datasets and shows summary statistics. In the table below, $S$ represents the size of the dataset and $D$ represents the dimension of the data.} \label{tab:cap1}
\begin{center}
\begin{tabular}{l|m{30pt}m{30pt}}
\hline
Dataset                & S       & D  \\ \hline
Boston Housing         & 506     & 13 \\
Wine Quality           & 1,599   & 11  \\
Forest Fires           & 517     & 13  \\
Concrete Compression Strength  & 1,030   & 9  \\
Energy Efficiency      & 768    & 8  \\
Naval Propulsion       & 11,934 & 16  \\
Combined Cycle Power Plant  & 9,568  & 4  \\
Protein Structure      & 45,730   & 9 \\
Online News Popularity & 39,797   & 61  \\ \hline
\end{tabular}
\end{center}
\end{table}

\begin{table*}[t]
\begin{center}
\caption{Performance of constructing PIs on nine UCI regression datasets. Quality assessment metrics is $PICP$ and $MPIW$, with best results in bold. Here, the best results are determined by the following rules. That is, if its $PICP$ $\geq$ 0.95, it is the best. If $PICP$ for all methods is less than 0.95, then the largest $PICP$ is the best. And, of all the methods, the one with the smallest $MPIW$ is the best. The average performance of all methods is also given below. } \label{tab:cap2}
\begin{tabular}{l|l|m{50pt}<{\centering} m{50pt}<{\centering} m{50pt}<{\centering} m{50pt}<{\centering}}
\hline
Datasets                                       & Metrics & $UBIP$        & $QR$          & $QR_{GBDT}$ & $IntPred$            \\ \hline
\multirow{2}{*}{Boston Housing}                & PICP    & \textbf{0.94} & 0.92          & 0.92   & 0.92                \\
                                               & MPIW    & \textbf{1.16} & 2.28          & 1.70   & 1.79                \\ \hline
\multirow{2}{*}{Wine Quality}                  & PICP    & \textbf{0.92} & 0.91          & 0.92   & 0.92                 \\
                                               & MPIW    & \textbf{2.22} & 2.64          & 2.47   & 2.54                 \\ \hline
\multirow{2}{*}{Forest Fires}                  & PICP    & 0.94          & \textbf{0.95} & 0.94   & 0.94                \\
                                               & MPIW    & \textbf{1.02} & 1.09          & 1.04   & 1.03                 \\ \hline
\multirow{2}{*}{Concrete Compression Strength} & PICP    & \textbf{0.94} & 0.94          & 0.94   & 0.94                 \\
                                               & MPIW    & \textbf{1.13} & 2.22          & 2.23   & 1.87                 \\ \hline
\multirow{2}{*}{Energy Efficiency}             & PICP    & \textbf{0.96} & 0.95          & 0.97   & 0.95                 \\
                                               & MPIW    & \textbf{0.67} & 1.73          & 0.79   & 1.56                 \\ \hline
\multirow{2}{*}{Naval Propulsion}              & PICP    & \textbf{0.98} & 0.98          & 0.98   & 0.98                 \\
                                               & MPIW    & \textbf{0.70} & 0.89          & 1.34   & 0.73                 \\ \hline
\multirow{2}{*}{Combined Cycle Power Plant}    & PICP    & \textbf{0.95} & 0.95          & 0.95   & 0.95                 \\
                                               & MPIW    & \textbf{0.83} & 0.97          & 0.90   & 0.84                 \\ \hline
\multirow{2}{*}{Protein Structure}             & PICP    & \textbf{0.95} & 0.95          & 0.95   & 0.95                 \\
                                               & MPIW    & 2.17          & 2.76          & 2.36   & \textbf{2.15}        \\ \hline
\multirow{2}{*}{Online News Popularity}        & PICP    & \textbf{0.95} & 0.95          & 0.96   & 0.95                 \\
                                               & MPIW    & \textbf{1.16} & 1.27          & 1.72   & 1.38                 \\ \hline
\multirow{2}{*}{Average Performance}           & PICP    & \textbf{0.95} & 0.94          & 0.95   & 0.94                 \\
                                               & MPIW    & \textbf{1.23} & 1.76          & 1.62   & 1.54                 \\ \hline
\end{tabular}
\end{center}
\end{table*}

\subsubsection{Ensemble Methodology}
Ensembles of neural networks have long been used to estimate predictive uncertainty ~\cite{heskes1997practical}~\cite{tibshirani1996comparison}. A group of neural networks are trained with different parameter initializations~\cite{lakshminarayanan2017simple} or with different sub-sampling training dataset. The variance or entropy of the predictions is interpreted as its epistemic uncertainty. Related algorithms include bootstrapping~\cite{johnson2001introduction}, and bagging from statistics~\cite{breiman1996bagging}, etc. Following the work~\cite{pearce2018high}, we have adopted a simple but effective ensemble principle: each individual model is a UBPI model initialized with different parameters. We assume that the size of the ensemble of neural networks is $m$ as shown in Figure \ref{fig3}. The ensemble PIs is the averaged PIs of all individual models,

\begin{equation}
  \mu_{\hat{y}_{Li}} = \frac{1}{m}\sum_{j = 1}^{m}\hat{y}_{Lij},
\end{equation}
\begin{equation}
  \mu_{\hat{y}_{Ui}} = \frac{1}{m}\sum_{j = 1}^{m}\hat{y}_{Uij}.
\end{equation}
And the variance is:
\begin{equation}
  \sigma^2_{\hat{y}_{Li}} = \frac{1}{m-1}\sum_{j=1}^{m}(\hat{y}_{Lij} - \mu_{\hat{y}_{Li}})^2,
\end{equation}
\begin{equation}
  \sigma^2_{\hat{y}_{Ui}} = \frac{1}{m-1}\sum_{j=1}^{m}(\hat{y}_{Uij} - \mu_{\hat{y}_{Ui}})^2.
\end{equation}

According to eq.(\ref{eq2}), the final PIs is:
\begin{equation}
  \tilde{y}_{Li} = \mu_{\hat{y}_{Li}} - \sigma^2_{\hat{y}_{Li}},
\end{equation}

\begin{equation}
  \tilde{y}_{Ui} = \mu_{\hat{y}_{Ui}} + \sigma^2_{\hat{y}_{Ui}},
\end{equation}
where $i$ is the index of $n$.
This method is scalable and easily implementable.

\section{Experiment}
In order to demonstrate the effectiveness of our method in detail, in this section we conducted a large number of experiments with our method on multiple toy datasets and real world datasets. Experimental results show that our method can capture the uncertainty well and construct the optimal prediction intervals.

\subsection{Regression on Toy Dataset}
In order to qualitatively evaluate the performance of the loss function we designed, the following formula is used to generate a one-dimensional toy regression dataset.
\begin{equation}
  y = 2cos(0.2x) + 0.2cos(10x) + 0.7cos(20x) + \epsilon,
\end{equation}
where $\epsilon \sim N(0, 0.1)$.

The results is shown in Figure \ref{fig1}. The optimal  prediction intervals should be as narrow as possible on the premise of achieving predefined confidence level $P_c$. Here, we set $P_c$ to 95\%. From the figure, we can see that the PIs predicted by our method tightly wraps the data points. And more than 95\% of the data points are in our PIs (the total number of data points is 100).

\subsection{Epistemic Uncertainty  Visualization}
To demonstrate the effectiveness of our method in capturing epistemic uncertainty, we generated some one-dimensional toy data for experiments. The function used for data point generation is:
\begin{equation}
  y = 1.5sin(x) + \epsilon,
\end{equation}
where $\epsilon \sim N(0, x^2)$.

The results is shown in Figure \ref{fig2}. From the figure, we can see that our method is very sensitive to the density of data distribution. The constructed PIs is in line with our expectation, that is, where there is less training data, the epistemic uncertainty increases, which is reflected in the increase of interval width. As we can see, where the data points are dense, the PIs is very narrow.

\subsection{Regression on Real World Datasets}
In our next experiment, we compare our method with several baseline methods for constructing PIs. The details of the experiment are described below.
\subsubsection{Experimental Setup and Baseline Methods}
The neural network structure we use is one hidden layer of 50 neurons with $ReLU$ activation. The exceptions to the experimental setup are the two largest datasets: $Online\,News\,Popularity$ and $Protein\,Structure$. For these two datasets, the network structure is one hidden layer of 100 neurons.

Several baseline methods are used for comparison, including Quantile Regression ($QR$)~\cite{koenker2001quantile}, Gradient Boosting Decision Tree with Quantile Loss ($QR_{GBDT}$) implemented in the Scikit-learn package~\cite{pedregosa2011scikit} and quality-based PI construction method $IntPred$~\cite{rosenfeld2018discriminative}. For all datasets, the ensemble size of neural networks is five.

\begin{table}[]
\begin{center}
\caption{The performance of different values of hyper-parameter $\lambda$ on $Protein\,Structure$ and $Online\,News$ $Popularity$ datasets} \label{tab:cap3}
\begin{tabular}{l|cc|cc}
\hline
\multirow{2}{*}{} & \multicolumn{2}{c|}{Protein Structure}                            & \multicolumn{2}{c}{Online News Popularity}          \\
                          & \multicolumn{1}{m{30pt}<{\centering}}{PICP} & \multicolumn{1}{m{30pt}<{\centering}|}{MPIW} & \multicolumn{1}{m{30pt}<{\centering}}{PICP} & \multicolumn{1}{m{30pt}<{\centering}}{MPIW} \\ \hline
$\lambda = 5$              & 0.947                         & 2.212                          & 0.925                         & 0.668                         \\
$\lambda = 10$             & 0.950                         & 2.271                          & 0.935                         & 0.761                         \\
$\lambda = 20$             & 0.953                         & 2.377                          & 0.948                         & 1.050                         \\
$\lambda = 30$             & 0.957                         & 2.510                          & 0.951                         & 1.155                         \\
$\lambda = 40$             & 0.965                         & 2.708                          & 0.953                         & 1.213                         \\
$\lambda = 50$             & 0.969                         & 2.907                          & 0.958                         & 1.462                         \\
$\lambda = 60$             & 0.973                         & 3.067                          & 0.960                         & 1.583                         \\ \hline
\end{tabular}
\end{center}
\end{table}

\subsubsection{Benchmark Datasets and Evaluation Metrics}
To measure the quality of PIs. We conduct experiments on nine UCI regression datasets as shown in Table \ref{tab:cap1}. The datasets are randomly divided into training set and test set, accounting for 90\% and 10\% respectively.

In this experiment, we used two quality assessment metrics: $PICP$ and $MPIW$. The results are shown in Table \ref{tab:cap2}. We repeated the experiments 20 times except for $Online\,News\,Popularity$ and $Protein\,Structure$. For these two datasets, the experiment is repeated five times. And the results are presented as averages.

\subsubsection{Performance Comparison and Discussion}
As we can see from Table \ref{tab:cap2}, our method achieves the best performance on most datasets. And overall, the average performance is better than all other methods. Average $PICP$ reached the predefined confidence level $P_c$, which is slightly better than other methods. At the same time, our method can obtain narrower PIs. Compared with $QR$ and $QR_{GBDT}$, our method can significantly reduce the PI width by 30.1\% and 24.1\% respectively. And compared with $IntPred$, the width was narrowed by 20.1\%.

\begin{figure} [t]
\centering
\subfigure[$Concrete\,Compression\,Strength$] { \label{fig:a}
\includegraphics[width=0.9\columnwidth]{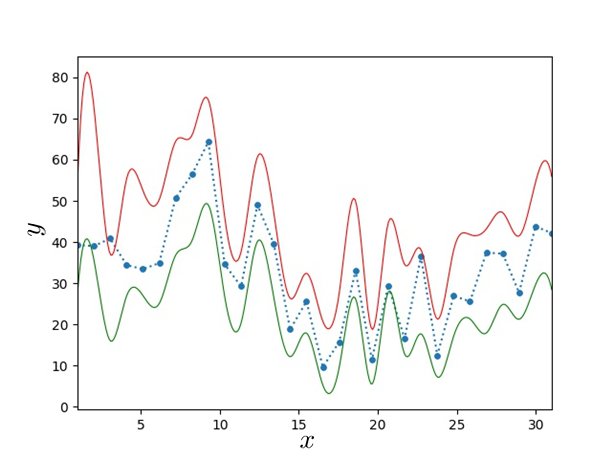}
}
\subfigure[$Combined\,Cycle\,Power\,Plant$] { \label{fig:b}
\includegraphics[width=0.9\columnwidth]{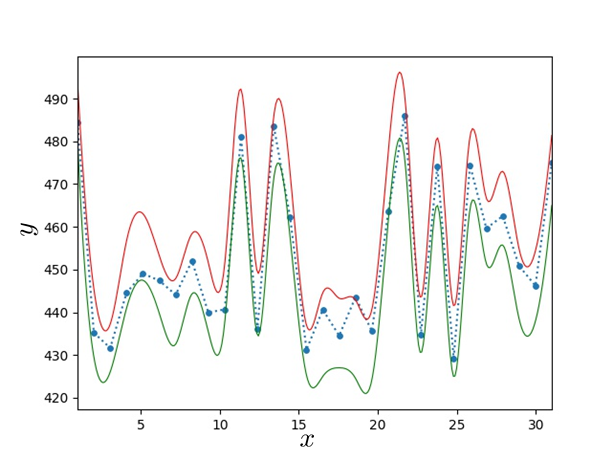}
}
\caption{Prediction intervals constructed on some test samples of $Concrete\,Compression\,Strength$ and $Combined\,Cycle\,Power\,Plant$. The $x$-axis denotes the index of test samples and $y$-axis denotes the value of the label. The blue dots are test data points. The red line and green line represent the upper bounds and lower bounds of the prediction intervals respectively.}
\label{fig4}
\end{figure}

At the same time, as shown in Figure \ref{fig4}, we construct and visualize the prediction intervals of some test samples. The effectiveness of our method can be seen intuitively from the diagram.

\subsubsection{The Effect of Different Values of Hyper-parameter $\lambda$}
In Table \ref{tab:cap3}, we show the effect of hyper-parameter $\lambda$ with different values. We can see $\lambda$ can controls the trade-off between $MPIW$ and $PCIP$. When we want the high coverage probability of PIs, the average PI width will increase. On the contrary, when we want the average PI width to be narrower, the coverage probability of PIs will decrease. This phenomenon is in line with our logic. In practical application, we can get the desired PIs coverage probability by adjusting $\lambda$.

\section{Conclusion And Future Works}
In this paper, the problem of estimating uncertainty associate with constructing optimal prediction intervals. We first give a comprehensive introduction to two kinds of uncertainties and the construction of prediction interval. And we give an unified literature review on uncertainty estimation. We estimate the uncertainty in deep learning from the perspective of prediction intervals. In general, we design a special loss function based on the theory of uncertainty estimation. We do not need uncertainty labels to learn uncertainty. Instead, we only need to supervise the learning of the regression task. We learn the aleatory uncertainty implicitly
from the loss function. The experiments show that our method can construct compact optimal prediction intervals. And it can capture the epistemic uncertainty sensitively.  In the area of sparse data distribution, the uncertainty increases obviously. In addition, the experimental results on some public regression datasets show the effectiveness of our method.

For future works, we believe that more innovative neural network ensembles method can be designed. In addition, there is a lack of quantitative evaluation metrics of model uncertainty currently, which should be a research direction to be paid attention to.


%



\ifCLASSOPTIONcompsoc
%
%

\ifCLASSOPTIONcaptionsoff
  \newpage
\fi



%
%
%
\bibliography{references}{}
\bibliographystyle{IEEEtran}
%
%
%
%
%




\end{document}